\documentclass{article}

\PassOptionsToPackage{numbers, compress}{natbib}

\usepackage[preprint]{airov26}

\usepackage[utf8]{inputenc} 
\usepackage[T1]{fontenc}    
\usepackage{booktabs}       
\usepackage{amsfonts}       
\usepackage{nicefrac}       
\usepackage{microtype}      
\usepackage{xcolor}         

\usepackage{amsmath}
\usepackage{cleveref}
\usepackage{graphicx}

\title{Flow Matching for Conditional MRI-CT and CBCT-CT Image Synthesis}

\author{%
  Arnela Hadzic, Simon Johannes Joham, Martin Urschler\\
  Institute for Medical Informatics, Statistics and Documentation\\
  Medical University of Graz\\
  \texttt{\{arnela.hadzic, martin.urschler\}@medunigraz.at} \\
}

\begin{document}

\maketitle

\begin{abstract}
Generating synthetic CT (sCT) from MRI or CBCT plays a crucial role in enabling MRI-only and CBCT-based adaptive radiotherapy, improving treatment precision while reducing patient radiation exposure. 
To address this task, we adopt a fully 3D Flow Matching (FM) framework, motivated by recent work demonstrating FM's efficiency in producing high-quality images. 
In our approach, a Gaussian noise volume is transformed into an sCT image by integrating a learned FM velocity field, conditioned on features extracted from the input MRI or CBCT using a lightweight 3D encoder.
We evaluated the method on the SynthRAD2025 Challenge benchmark, training separate models for MRI $\rightarrow$ sCT and CBCT $\rightarrow$ sCT across three anatomical regions: abdomen, head and neck, and thorax. 
Validation and testing were performed through the challenge submission system. 
The results indicate that the method accurately reconstructs global anatomical structures; however, preservation of fine details was limited, primarily due to the relatively low training resolution imposed by memory and runtime constraints. 
Future work will explore patch-based training and latent-space flow models to improve resolution and local structural fidelity.
\end{abstract}

\section{Introduction}
Accurate dose calculation in radiotherapy (RT) requires quantitative tissue information typically obtained from computed tomography (CT). However, CT exposes patients to ionizing radiation, offers limited soft-tissue contrast, and is often unavailable in treatment rooms. Magnetic resonance imaging (MRI) provides superior soft-tissue contrast without radiation but lacks attenuation information, while cone-beam CT (CBCT) offers in-room imaging and lower radiation doses but suffers from artifacts. 

Converting MRI or CBCT to synthetic CT (sCT) can enable MRI-only workflows, adaptive MRI-based RT, and CBCT-based adaptive RT, improving precision while reducing radiation exposure. 
Despite rapid methodological progress, the lack of public datasets and standardized evaluation of CT synthesis approaches hampers objective comparison. 
The SynthRAD2025 Challenge addresses this gap by providing a public benchmark for MRI $\rightarrow$ sCT and CBCT $\rightarrow$ sCT methods with unified data, metrics, and evaluation protocols. Compared to the SynthRAD2023 Challenge \cite{huijben2024generating}, which focused on a smaller brain and pelvis dataset, the SynthRAD2025 Challenge includes over 2,300 cases across abdomen (AB), head and neck (HN), and thorax (TH) regions from multiple centers, providing a larger and more diverse benchmark for evaluating sCT generation methods.

To tackle the sCT generation problem for the SynthRAD2025 Challenge, we employ a Flow Matching (FM) generative model \cite{lipman2022flow,liu2022flow}. FM offers an efficient way to model complex data distributions by learning continuous probability flows from Gaussian noise to target images. 
Compared to generative adversarial networks traditionally used in medical image synthesis \cite{nie2017medical,neff2017generative}, FM provides more stable training and distribution coverage. Moreover, it requires fewer integration steps than diffusion models \cite{muller2023multimodal,dorjsembe2024conditional,hadzic2024synthetic}, thereby achieving faster inference while maintaining high image quality \cite{zhang2024flow,zhang2024mutli,bogensperger2025flowsdf,hadzic2025flow}.\\
In this work, we adapt the FM formulation \cite{lipman2022flow} to explicitly condition the model on the given MRI/CBCT volumes using a lightweight 3D encoder. To the best of our knowledge, this is the first application of Flow Matching to conditional synthetic CT generation.

\section{Materials and Methods}
\subsection{Dataset}\label{sec:material}
The dataset provided by the SynthRAD2025 Challenge \cite{thummerer2025synthrad2025} contains paired MR-CT images for Task 1 (MRI $\rightarrow$ sCT) and CBCT-CT images for Task 2 (CBCT $\rightarrow$ sCT), acquired from multiple centers with varying sizes. For Task 1, image dimensions range from 238-608 $\times$ 239-553 $\times$ 42-164, while for Task 2, image dimensions range from 265-560 $\times$ 225-560 $\times$ 49-139 voxels across anatomical regions.
All images have an anisotropic spacing of $1\times 1 \times 3\ \text{mm}^3$.

The Task 1 training set consisted of 578 paired MRI-CT volumes, comprising 175 AB, 221 HN, and 182 TH cases. An additional 89 unpaired MRI volumes were provided for validation. For Task 2, the training set included 1,472 paired CBCT-CT volumes (309 AB, 325 HN, and 321 TH), and 148 unpaired CBCT volumes were provided for validation. The test phases for both tasks were conducted via the challenge platform on 223 MRI volumes (Task 1) and 369 CBCT volumes (Task 2). Since the ground-truth CT images for the validation and test sets were not accessible to participants, we performed an internal data split using 75\% of the training data for training and 25\% for validation. The setup that achieved the lowest MSE on this internal validation set was chosen and the final models were trained from scratch on all available training images.

\subsection{Methodology}
We performed CT synthesis from MRI or CBCT, respectively, using a conditional Flow Matching framework \cite{lipman2022flow,liu2022flow}. 
In our formulation, the target CT image $x_1 \sim p_1$ is connected to a base Gaussian noise sample $x_0 \sim p_0 = \mathcal{N}(0, I)$ through a linear interpolation path 
\begin{equation*}
x_t = t\, x_1 + \sigma_t \, \epsilon,
\quad \epsilon \sim \mathcal{N}(0, I),
\quad \sigma_t = 1 - (1 - \sigma_{\min})\, t,
\end{equation*}
with $\sigma_{\min} = 10^{-5}$ \cite{lipman2022flow}.
For this path, the corresponding probability flow velocity field is given by
\begin{equation*}
u_t(x_t, t \mid x_1)
=
\frac{x_1 - (1 - \sigma_{\min})\, x_t}{1 - (1 - \sigma_{\min})\, t}.
\end{equation*}

We train the Flow Matching model on paired MRI/CT or CBCT/CT images. Specifically, we draw $t \sim \mathcal{U}(0,1)$, construct $x_t$ from the target CT, and train the network $v_\theta(x_t, t \mid c)$ to approximate $u_t$, where $c$ is the corresponding MRI/CBCT conditioning image. The conditioning volume is processed by a lightweight 3D convolutional encoder, and its features are concatenated with $x_t$ before being passed to a 3D U-Net that predicts a velocity field. The objective combines L1 and MSE losses:
\begin{equation*}
\mathcal{L}(\theta) =
\lambda_{\mathrm{L1}} \| v_\theta(x_t, t \mid c) - u_t \|_1 +
\lambda_{\mathrm{MSE}} \| v_\theta(x_t, t \mid c) - u_t \|_2^2 ,
\end{equation*}
with $\lambda_{\mathrm{L1}} = \lambda_{\mathrm{MSE}} = 1$. 

At inference time, we sample $x_0 \sim \mathcal{N}(0, I)$ and solve the probability-flow ODE
\begin{equation*}
\dot{x}_t = v_\theta(x_t, t \mid c), \quad t : 0 \to 1,
\end{equation*}
where $v_\theta$ is the learned velocity field. Integration is performed with a 4th-order Runge-Kutta (RK4) solver using 32 steps. Starting from random noise $x_0$, the sample is gradually transformed along the learned flow to produce the final synthetic CT image $x_1$.

For each task and anatomical region, separate models were trained using the corresponding paired MRI-CT or CBCT-CT training images (see \Cref{sec:material}).

\section{Experimental Setup}
\subsection{Implementation Details}
\subsubsection{Data preprocessing} 
All images were resampled to a uniform spatial resolution of $128 \times 128 \times 128$ voxels with an isotropic spacing of $1 \times 1 \times 1\ \text{mm}^3$. Intensity normalization was applied independently to each modality: MR images underwent z-score normalization followed by clipping to the range $[-3,\ 3]$, while CBCT and CT images were clipped to the Hounsfield Unit (HU) range $[-1024,\ 3071]$ and scaled by dividing by 1000, resulting in a normalized range of $[-1024,\ 3071]$. 

\subsubsection{Architectural and training details} 
To predict velocity fields, we employed a 3D conditional U-Net architecture from \cite{dorjsembe2024conditional}. We adapted the network to first process the conditioning MR or CBCT image through two $3\times3\times3$ convolutional layers with ReLU activations, producing feature maps with 64 channels. These features were then concatenated with the input CT image and passed to the main 3D U-Net \cite{dorjsembe2024conditional}. The U-Net consisted of four resolution levels, each containing one residual block. The number of convolutional channels at each level was determined by a set of channel multipliers $(1, 1, 2, 3, 4)$, scaled from a base of 64 channels. This resulted in 64, 64, 128, 192, and 256 channels across the levels. Self-attention mechanisms were applied at resolutions corresponding to $16 \times 16 \times 16$ and $8 \times 8 \times 8$ feature maps. All residual blocks used dropout with a probability of 0.05.

Training data was augmented with random 3D translations ($\pm$ 5 voxels) and rotations ($\pm 0.1$ radians). The model was optimized using AdamW with a learning rate of $10^{-4}$ and a weight decay of $10^{-5}$, with an effective batch size of 2. This configuration was used for each task (MRI $\rightarrow$ sCT, CBCT $\rightarrow$ sCT) and anatomical region (AB, HN, TH), resulting in six models in total.

MRI $\rightarrow$ sCT models were trained for 100,000 steps, whereas CBCT $\rightarrow$ sCT models were trained for 50,000 steps. Implementation was done in PyTorch. The ODE solver (i.e., RK4) was implemented using the torchdiffeq library. Additional packages included SimpleITK, NumPy, and scikit-image. Training was performed on an 80 GB NVIDIA A100 and required approximately 3 days and 15 hours for MRI $\rightarrow$ sCT (HN), and 2 days and 9 hours for  CBCT $\rightarrow$ sCT (HN). Inference per scan took approximately 2 minutes on a 24 GB NVIDIA GeForce RTX 3090 GPU. The total number of trainable parameters was 41,237,057.

\subsubsection{Data postprocessing} 
Synthesized CT images were resampled back to the original spatial resolution of the input MR or CBCT image using nearest-neighbor interpolation. Additionally, the original image spacing and origin were reapplied and intensity values were denormalized by multiplying by 1000 to recover values in the HU range $[-1024,\ 3071]$.

\subsection{Evaluation}\label{sec:evaluation}
Quantitative scores for our models on the validation set provided by the SynthRAD2025 Challenge were obtained via the submission system. The evaluation metrics used assessed both image similarity and geometric consistency. Image similarity between synthetic CT and target CT within the provided body mask was assessed using mean absolute error (MAE), peak signal-to-noise ratio (PSNR), and multi-scale structural similarity index (MS-SSIM). Geometric consistency was evaluated through automated segmentation (using TotalSegmentator \cite{wasserthal2023totalsegmentator} and nnUNet \cite{isensee2021nnu}), with performance quantified using the multiclass Dice coefficient (mDice) and the 95th percentile Hausdorff distance (HD95), averaged across all anatomical structures (abdomen, head and neck, and thorax).

\section{Results}
\begin{figure}[t]
\centering
\includegraphics[width=0.92\textwidth]{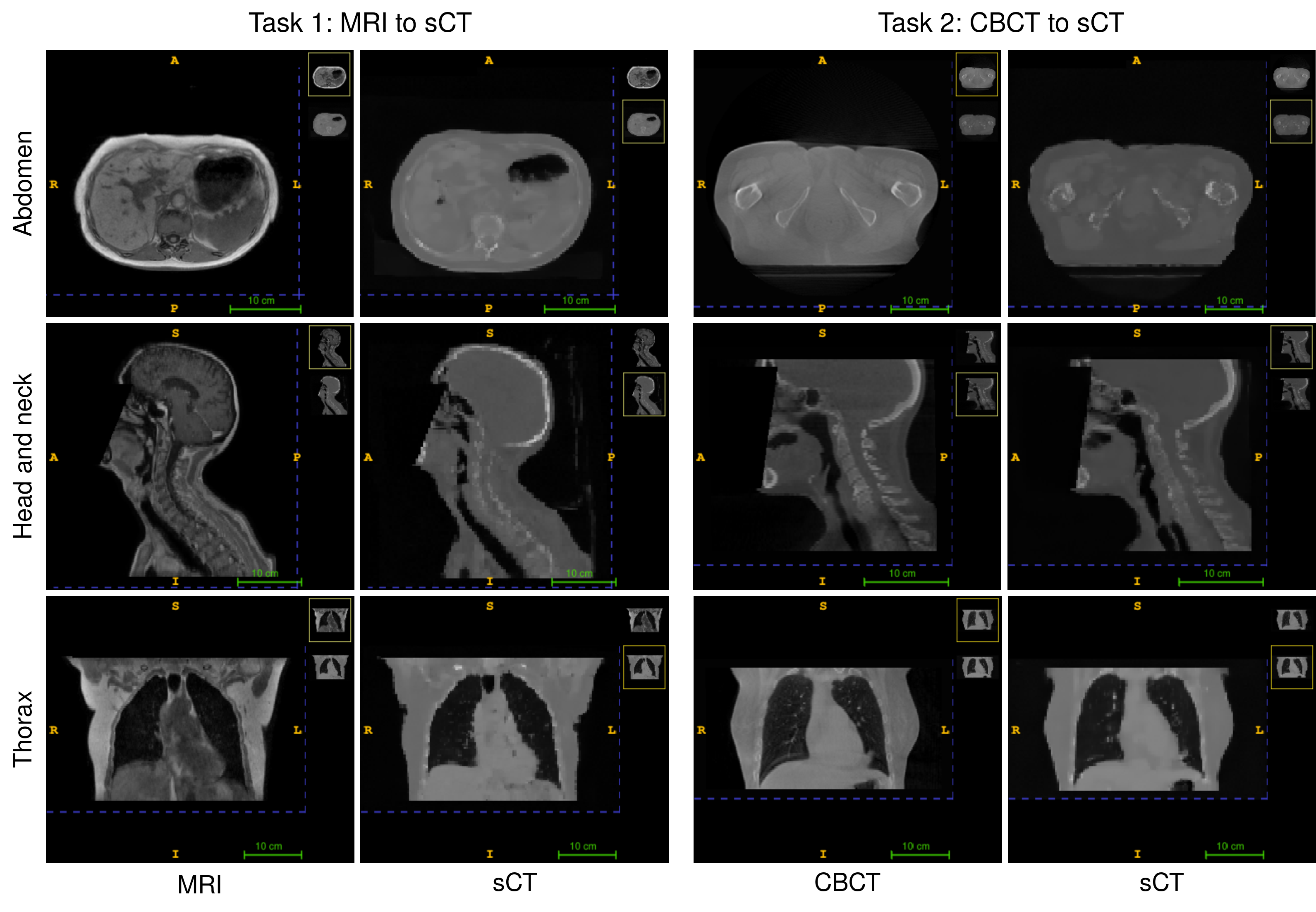}
\caption{Examples of synthetic CT images generated by our MRI $\rightarrow$ sCT (left column) and CBCT $\rightarrow$ sCT models (right column) for different anatomical regions.} \label{fig:results}
\end{figure}

\Cref{tab:results} presents the quantitative results for the MRI $\rightarrow$ sCT model on 89 validation images and for the CBCT $\rightarrow$ sCT model on 148 CBCT validation images. The results are reported using the metrics described in \Cref{sec:evaluation}, with both the mean and standard deviation (mean $\pm$ std) for each metric. Qualitative results are shown in \Cref{fig:results}, where examples of synthetic CT images for both tasks and all three anatomical regions are displayed. 

We observed no artifacts in the generated images, but they appear blurry due to the relatively low training resolution. While the global anatomy is captured well, the preservation of local details could be improved. This is also reflected in the quantitative scores reported in \Cref{tab:results}.

Overall, our approach was ranked 12th in the final SynthRAD2025 Challenge leaderboard. While our quantitative performance was lower than the top-ranked methods, such as the winning team which achieved an MAE of approximately 65 HU for Task 1, it is important to note the fundamental difference in methodology. The majority of competing participants, including the highest-ranking teams, utilized established supervised learning frameworks (e.g., highly optimized 3D U-Nets or ensembled CNNs) designed to learn the synthesis mapping directly in a discriminative manner. In contrast, our work follows a different approach by employing a generative probability flow framework. Although this generative approach faced challenges with local structural fidelity due to current resolution constraints stemming from GPU memory limitations, it introduces a generative modeling alternative that can be further scaled to capture complex data distributions.

\begin{table}[h]
\centering
\caption{Quantitative results obtained on the corresponding validation sets for both tasks.}\label{tab1}
    \begin{tabular}{l|c|c|c|c|c}
        \textbf{Model} &  MAE & PSNR & MS-SSIM & DICE & HD95\\
        \hline
        MRI $\rightarrow$ sCT & $146.17 \pm 27.91$ & $24.62 \pm 1.43$ & $0.82 \pm 0.08$ & $0.44 \pm 0.14$ & $18.54 \pm 10.70$\\ \hline
        CBCT $\rightarrow$ sCT & $114.74 \pm 23.12$ & $26.30 \pm 1.61$ & $0.88 \pm 0.04$ & $0.58 \pm 0.14$ & $12.88 \pm \phantom{0}7.73$\\
    \end{tabular}
\label{tab:results}
\end{table}

\section{Discussion and Conclusions}
In this work, we addressed the MR/CBCT $\rightarrow$ sCT task with a Flow Matching generative model conditioned on MR/CBCT inputs. The method consistently reproduced global anatomy (see \Cref{fig:results}), but preservation of fine structures remains suboptimal. We attribute this primarily to the training resolution ($128\times128\times128$) being lower than the original resolution of many scans. Scaling to higher resolutions was limited by memory and runtime constraints. Importantly, the same architecture and training setup were used for all models, without task- or anatomy-specific tuning.

In future work, we plan to adopt 3D patch-based training and inference, and explore latent-space flow models to enable higher effective resolution and improved local detail.

\begin{ack}

This research was funded in whole or in part by the Austrian Science Fund (FWF) 10.55776/PAT1748423.



\end{ack}

\bibliographystyle{unsrt}
\bibliography{paper}

\end{document}